\documentclass[journal]{IEEEtran}
\usepackage{graphicx}
\usepackage{times}
\usepackage{helvet}
\usepackage{courier}
\usepackage{amsmath} 
\usepackage{amssymb}  
\usepackage{amsthm} 
\usepackage{alltt}
\usepackage{xspace}
\usepackage{url}
\usepackage[table]{xcolor}
\usepackage{color,soul}

\begin{document}

\title{AI Challenges in Human-Robot Cognitive Teaming}

\author{
{\Large Tathagata Chakraborti$^1$, Subbarao Kambhampati$^1$, Matthias Scheutz$^2$, Yu Zhang$^1$}\\[1ex]
$^1$ Department of Computer Science, Arizona State University, Tempe, AZ 85281 USA\\
{\tt \{ tchakra2, rao, yzhan442 \} @ asu.edu}\\[1ex]
$^2$ Department of Computer Science, Tufts University, Medford, MA 02155 USA\\
{\tt matthias.scheutz@tufts.edu} \vspace{-7pt}
}

\maketitle


\begin{abstract}

Among the many anticipated roles for robots in the future is that of being a
human teammate. Aside from all the technological hurdles that
have to be overcome with respect to hardware and control to make robots
fit to work with humans, the added complication here is that
humans have many conscious and subconscious expectations of their
teammates -- indeed, we argue that teaming is mostly a {\em cognitive} 
rather than {\em physical} coordination activity. 
This introduces new 
challenges for the AI and robotics community and requires fundamental 
changes to the traditional approach to the design of autonomy.
With this in mind, we propose an update to the 
classical view of the intelligent agent architecture, highlighting
the requirements for mental modeling of the
human in the deliberative process of the autonomous agent. 
In this article, we outline briefly the recent efforts of ours, and others in the community, towards developing cognitive teammates along these guidelines.

\end{abstract}



\section{Introduction}
\label{sec:intro}

An increasing number of applications demand that humans and robots
work together.  
Although a few of these applications can be handled
through ``teleoperation'', technologies that act in concert with the
humans in a teaming relationship with increasing levels of autonomy
are often desirable if not required.  
Even with a sufficiently robust human-robot
interface, robots will still need to exhibit
characteristics common in human-human teams in order to be good team players.
This includes the ability to
recognize the intentions of the human teammates, and to interact in
a way that is comprehensible and relevant to them - 
autonomous robots need to understand and adapt to human behavior 
in an efficient manner, much like humans adapt to the
behavior of other humans.
Humans are often able to produce such teaming behavior proactively 
due to their ability (developed through centuries of evolution of 
a variety of implicit or explicit visual, auditory and 
contextual cues) to quickly 
$(1)$ recognize teaming context in terms of the current status of the team
task and states of the teammates,
$(2)$ anticipate next team behavior under the current context 
to decide individual subgoals to be achieved for the team, and 
$(3)$ take proper actions to support the advancement of those subgoals
with the consideration of the other teammates.

The three steps above form a tightly coupled integrated loop 
during the coordination process, which
is constantly evolving during teaming experience. 
Critically, humans will likely expect all of the above capabilities from a robotic
teammate, as otherwise team dynamics will suffer.
As such, the challenge in human-robot teaming is primarily {\em cognitive}, 
rather than physical.
Cognitive teaming allows the robots to adapt more proactively to the 
many conscious and subconscious expectations of their human teammates. 
At the same time, improper design
of such robot autonomy could increase the human's cognitive load, 
leading to the loss of teaming situation awareness, 
misaligned coordination, 
poorly calibrated trust, 
and ultimately slow decision making, deteriorated teaming performance, 
and even safety risks to the humans. 
As designers of robotic control architectures, 
we thus have to first isolate the necessary functional capabilities that are common 
to realizing such autonomy for teaming robots.
The aim of the article is to do
just that and thus provide a framework that can serve as the basis for
the development of cognitive robotic teammates.

\section{Related Work}

\begin{figure*}
\centering
\includegraphics[width=\textwidth]{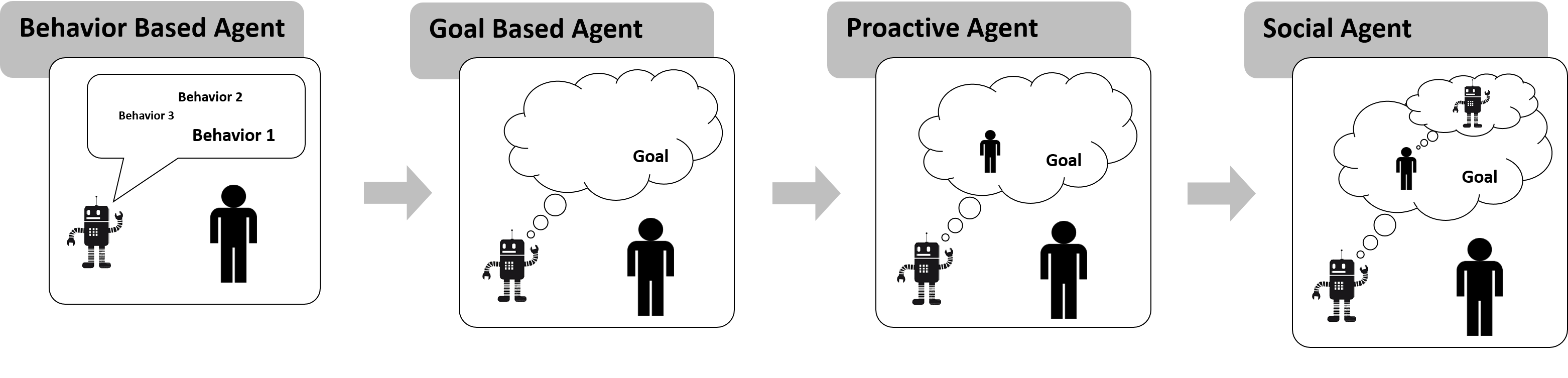}
\caption{An categorical view of different types of agents. Each type is deeper than the previous types in terms of modeling complexity (from left to right).
}
\label{fig:complexity}
\end{figure*}

In human-human teams, 
it is well understood that every team member maintains a cognitive model of the other teammates they interact with \cite{nancy2013}. 
These models not only captures their physical states,
but also mental states such as the teammate intentions and preferences,
which can significantly influence how an agent interacts with the 
other agents in the team. 
Although such modeling has been identified as an important characteristic of effective teaming \cite{nancy2015, nancy2015-2, nancy-new},  
it is less clear how it is maintained at the individual level.  
Furthermore, the relative importance of different aspects of 
such models cannot be easily isolated 
in experiments with human teammates, 
but must be separately considered for robots 
since they often require very different modeling technologies. 
Such forms of modeling 
allows the robots to understand their human partners,
and in turn use this knowledge to plan their coordination to improve teaming experience. 
However, although there exists work 
that has investigated the various aspects of this modeling \cite{proactive, 6942970, zhang2016plan},
a systematic summary of the important challenges is still missing. 

Next, we provide a review of the related work in terms of agent types (Figure \ref{fig:complexity})
that can be used to implement robotic teammates,
following the categorization in \cite{Wooldridge:2009}.
The first two types correspond to classical agent architectures in robotics and artificial intelligence. 
We list them to facilitate the comparison of earlier teaming agents and cognitive teaming agents. 
We show that all of them fall within a spectrum of agents
that differ based on the extent to which the agent interaction with the external world and other agents is modeled.

\subsection{Behavior-based agent} 
Behavior-based agents
\cite{Brooks91intelligencewithout} have been an important design
paradigm for embodied agents (especially robots), in which complex
behaviors result from a collection of basic behaviors interacting with each other.
 These basic behaviors often operate in parallel via cooperative or competitive
arbitration schemes \cite{scheutzandronache04cybernetics}.
Behavior-based agents have been applied to various tasks such as
formation control \cite{arkin2,balch}, box pushing
\cite{parker,gerkey2}, navigation \cite{parker2}, and surveillance
\cite{yu3}.  One issue with behavior-based agent is that the
interactions between basic behaviors often have to be provided
manually.  This can quickly become impractical when complex
interactions are desired.  Furthermore, since this type of agent does
not maintain a model of the world, 
it cannot reason about its dynamics and hence is often purely reactive.  

\subsection{Goal-based agent}
In contrast to behavior-based agents, a
goal-based agent maintains a model of the world (Fig.~\ref{fig:agent})
and how its actions can change the state of the world.  
As a result, it can predict how the
world will respond before it executes any action.  The
earliest agent of this type is Shakey \cite{shakey}.  The
model that is maintained is often specified at a factored level using planning languages, 
such as STRIPS \cite{STRIPS72}, PDDL \cite{fox2003pep} 
or its probabilistic extensions \cite{Sanner_relationaldynamic}, 
or at the atomic level using MDP specification \cite{Puterman:1994, proactive}.  
A goal-based agent
can also maintain its own epistemic state \cite{epistemic}, such as beliefs and desires
\cite{Rao95bdiagents, Georgeff1999}.  
Goal-based agents
typically assume that the given model is complete, 
which may not be realistic
in open world domains \cite{acm-tist}.
	
Both behavior and goal based agent can handle multi-agent coordination \cite{yu3, parker2, arkin2, Nissim, Grosz}. 
However, it is often assumed that the team is given a specific goal,
and the team members either are provided information about each other a priori,
or can explicitly exchange such information. 
As a result, agents can readily maintain a model of the others in teaming.
While this assumption may be true for robots teaming with robots,
we would definitely not observe such convenience in human-robot teams
(e.g., requiring humans to provide the information constantly can significantly increase their cognitive load);
furthermore, the goal is often spontaneous rather than given.
As a result, a robotic teammate that is solely
behavior or goal based can only handle specific tasks 
and will rely on human inputs for task assignments. 

\subsection{Proactive agent}	
A proactive agent, on the other hand,
is supposed to maintain a model of the others through both observations
and communication (if available and necessary).
This model is not only about the others' physical state (e.g., location), 
but also mental state which includes their goals \cite{kartik-iros-2014},
capabilities \cite{yu-aamas} (which include the consideration of physical capabilities), 
preferences \cite{nguyen-partialp-2012}, 
and knowledge \cite{chitta-rico}.  

Given that none of these are directly available,
they must be inferred  \cite{geffner-pr1}
or learned \cite{yu-aamas,Abbeel,irl-trace} from observations.
As a result, the model of the other agents
is often subject to a high-level of incompleteness and
uncertainty.
This is especially true when human teammates are involved.  
Nevertheless, 
even such an {\it approximate} model of other agents can be important for efficient teaming when used properly.
For example, it can be used by the agents to plan for coordination
to exploit
opportunities to help the humans in a proactive way
\cite{tathagata-serendipity-iros, proactive} while avoiding conflicts
\cite{cirillo, tathagata-conflict-aamas}.

In addition to using the model of the others to plan for coordination,
a proactive agent can also act proactively to change
the others' modeling of itself when necessary \cite{chitta-rico}. 
For example, a robot can
explicitly convey its intention through natural languages
\cite{Tellex14askingfor} or gestures \cite{perzanowskietal98}
to let the human understand its intention to help or request for help.

\subsection{Social agent}
A deeper level of modeling is not only about the other agents,
but also the other agents' modeling
of the agent itself \cite{zhang2016plan}.  
This includes, for example, the others' expectation and trust of the agent itself. 
Such modeling allows the
robot, for example, to infer the human expectation of its own behavior 
and in turn choose behaviors that are consistent with this expectation.
Expectation and trust, in particular, represent the social aspects of agent interactions 
since they are particularly relevant when agents form groups or teams together. 
An agent that behaves socially \cite{mainprice2010planning, Dragan-RSS-13,zhang2016plan} allows the other agents to 
better understand and anticipate its behavior,
thus contributing to the maintenance of teaming situation awareness \cite{nancy2015}.
In human-human teams, social behaviors contribute significantly to fluent teaming \cite{scheutzetalforthcoming}.

Similar to a proactive agent, a social agent often has to learn and maintain
a model about the various social aspects from observations \cite{zhang2016plan}.
In addition to using these social aspects to guide its behavior generation, 
a social agent can also act to change
these aspects (via informing the others about the discrepancies in their modeling about itself 
{\em and} updating the same in its model of the others). 
For example, a robot can maintain its trust from the human
by constructing excuses when a task cannot be achieved \cite{excuses-icaps}
to provide explanations to the human from the robot's own perspective,
while taking into account the human's understanding of itself.

Although various types of agents can be used to realize a robotic teammate,
based on the above discussion,
the challenges that are introduced by the humans in the loop lie in particular 
in the implementation of a proactive and social agents. 
A common characteristics among these two types of agents is that 
both require the notion of {\em{\textbf{mental modeling}}} of the other teammates,
which cannot be directly observed and must be inferred cognitively. 
This is the key requirement of a cognitive teaming capability.  




\section{Transitioning to a Cognitive Teaming Agent} 
\label{sec:new_view}

\begin{figure}
\centering
\includegraphics[width=\columnwidth]{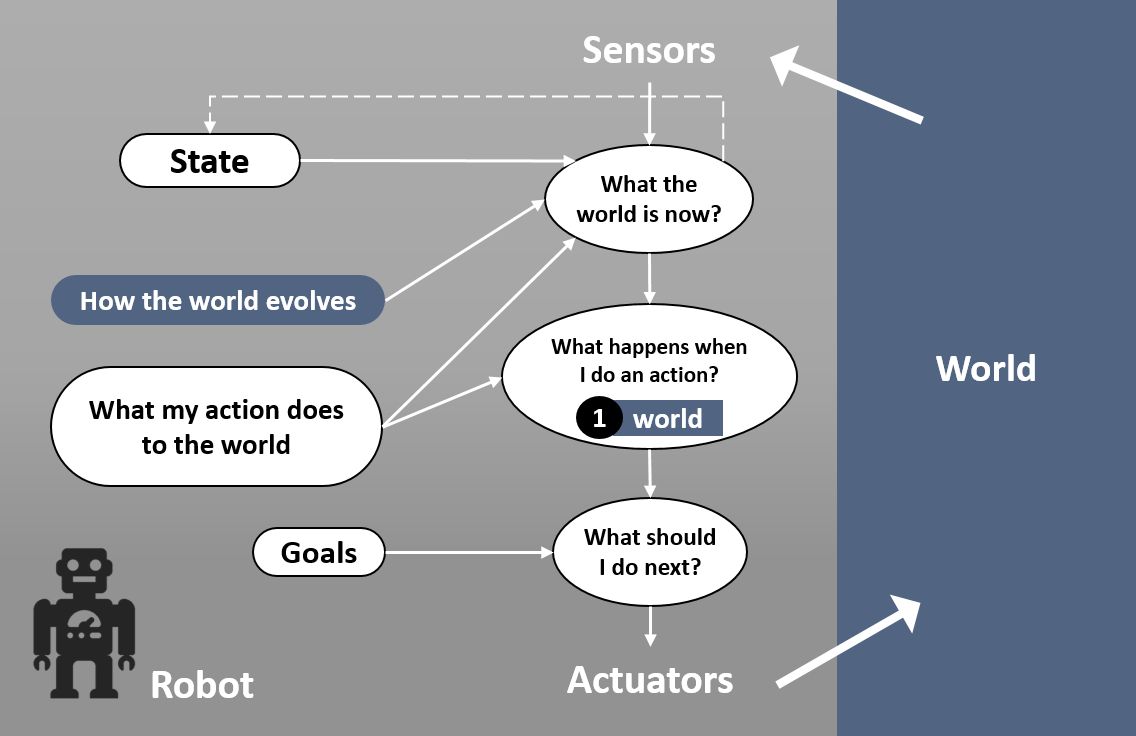}
\caption{Traditional view of the goal-based intelligent agent architecture \cite{russell_artificial_2010} that describes how the agent models the world, senses changes in the environment, plans to achieve goals and acts to execute the plans.}
\label{fig:agent}
\end{figure}

\begin{figure}
\centering
\includegraphics[width=\columnwidth]{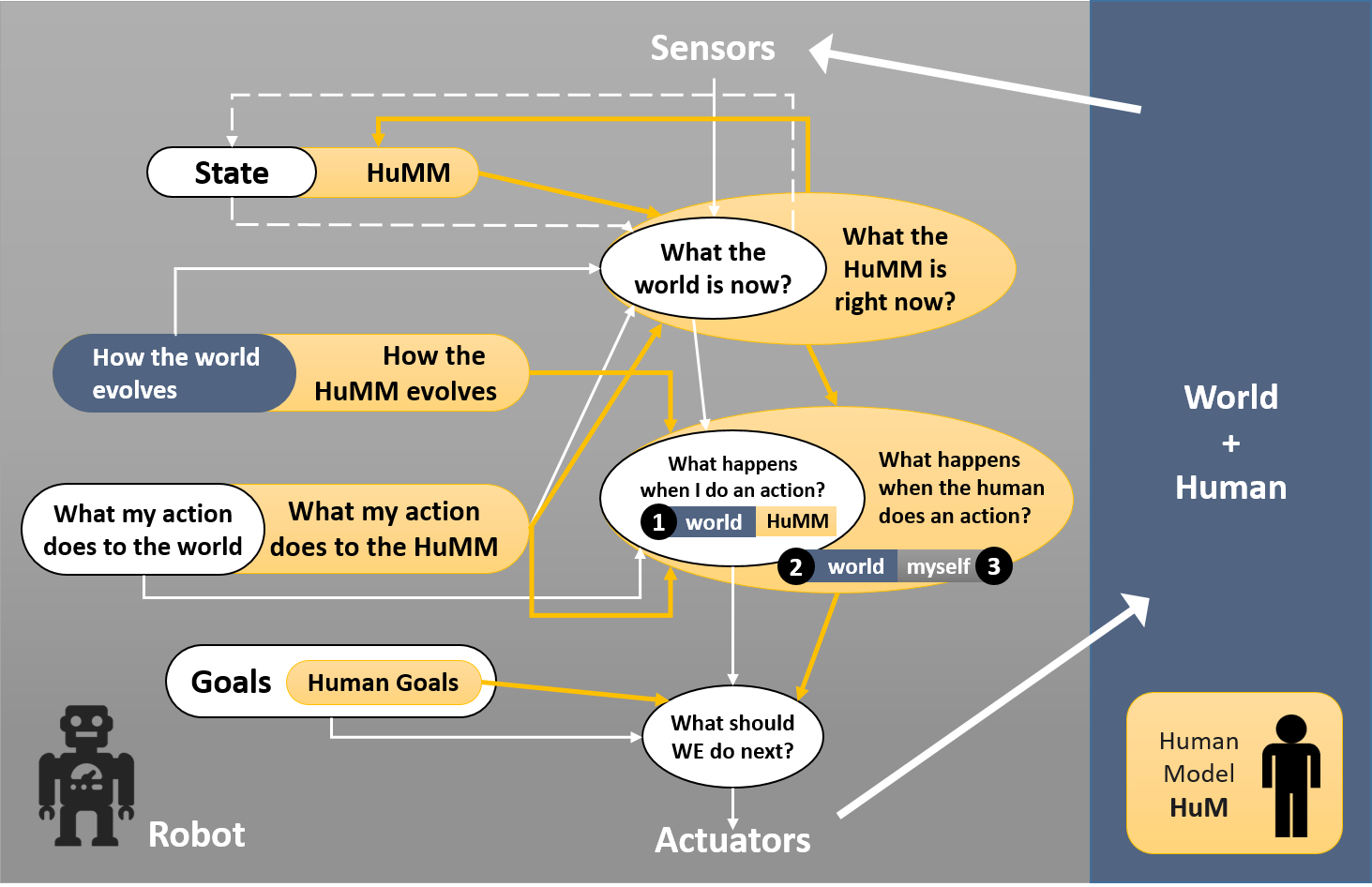}
\caption{An updated view of the architecture of a cognitive teaming agent acknowledging the need to account for the human's mental state by means of what we refer to as \textbf{Hu}man \textbf{M}ental \textbf{M}odeling or HuMM.}
\label{fig:newagent}
\end{figure}

In this section, we characterize how each step in the 
``{\em Sense-Model-Plan-Act}'' ({\bf SMPA}) cycle of the classical goal-based agent view in \cite{russell_artificial_2010} (shown in Figure \ref{fig:agent}) 
has to be updated to facilitate the mental modeling of the human in the loop 
in order to enable a truly cognitive teaming agent (shown in Figure \ref{fig:newagent}).
Specifically, we introduce the {\em Human Model} ({\bf HuM}) and the {\em Human Mental Model} ({\bf HuMM}) as key components in the agent's deliberative process.
Changes to {\em Model} in Figure \ref{fig:newagent} is a direct result of the requirement of human mental modeling. 
Coarsely speaking, changes to {\em Sense} contribute to the recognition of teaming context,
changes to {\em Plan} contribute to the anticipation of team behavior and {\em Act},
and changes to {\em Act} contribute to the determination of proper actions at both the action and motion levels.
In practice, these four functionalities are tightly integrated in the behavior loop.

\vspace{5pt}
\noindent {\textbf{Sense --}} 
The agent can no longer sense passively to
check that the preconditions of an action are satisfied, or after it
applies an action to the world to confirm that it is updated
accordingly ({\em ``what the world is like now''} in Figure \ref{fig:agent}).
In teaming scenarios, the agent
needs to proactively make complex sensing plans that interact
closely with other functionalities -- {\bf Model} and {\bf Plan} -- 
to maintain the correct mental state (such as intentions,
knowledge and beliefs) of its human teammates in order to infer
their needs. For example, how the robot should behave is
dependent on how much and what type of help the human requires,
which in turn depends on the observations about the human teammates
such as their behavior and workload.  Furthermore, the inference
about the human mental state should be informed by the human model
that the robot maintains about the human's capabilities and preferences.
Note that directly asking humans (i.e. explicit communication) 
is a specific form of sensing.
  
\vspace{5pt}
\noindent {\textbf{Model --}} 
Correspondingly, the state, i.e. {\em ``what the world is like now''},
needs to include not only environmental states, but also mental
states of the team members which
may not only include cognitive and affective states such as
the human's task-relevant beliefs, goals, preferences, and
intentions, but also, more generally, emotions, workload,
expectations, trust and etc.  
{\em ``What my actions do to the world''} then
needs to include the effects of the robot's actions on the team
member's mental state, in addition to the effects on their
physiological and physical states and the observable environment;
{\em ``How the world evolves''} now also requires rules that govern the
evolution of agent mental states based on their interactions with
the world (including information exchange through communication);
{\em ``What it will be like''} will thus be an updated state representation
that not only captures the world state, agent physiological and
physical state changes based on their actions and current states,
but also those mental state changes caused by the agent itself and
other team members.
	
\vspace{5pt}
\noindent {\textbf{Plan --}} 
{\em ``What action I should do''} now involves more
complex decision-making that must again also consider human mental state.
Furthermore, since the robot actions now can influence not only the
state of the world but also the mental state of the humans, the
planning process must also consider how the actions may influence
their mental state and even how to affect/manipulate such mental state.
For example, in teaming scenarios, it is important to maintain a
shared mental state between the teammates.  
This may require the robots
to generate behavior that is expected or predictable to the human
teammates such that they would be able to understand the robot's
intention.  
This can, in fact, be considered an implicit form of signaling or communication.
On the other hand, a shared mental state does not
necessarily mean that every piece of information needs to be
synchronized.
Given the limitation on human cognitive load, sharing
only necessary information is more practical between different
teammates working on different parts of the team task.  
A properly maintained shared mental state between the teammates can contribute
significantly to the efficiency of teaming since it can reduce the
necessity of explicit communication.

\vspace{5pt}
\noindent {\textbf{Act --}} 
In addition to physical actions, we now also have
communication actions that can change the mental state of the humans
by changing their beliefs, intents, etc.  
Actions to affect the human's 
mental state do not have to be linguistic (direct);
stigmergic actions to instrument the environment 
can also inform the humans such that
their mental states can be changed.  Given that an action plan is
eventually realized via the activation of effectors by providing
motor commands, {\it Act} must be tightly integrated with {\it
Plan}.  While {\it Plan} generates the sequence of actions to be
realized, motor commands can create different motion trajectories to
implement each action and can in turn impact how the plan would be
interpreted since different realizations can exert different
influences on the human's mental states based on the context.

\subsection*{An Exemplary Human-Robot Teaming Scenario} 
\label{sec:case_studies}

To better illustrate how mental modeling of teammates 
can contribute to the different capabilities needed for cognitive
teaming agents,
we will now consider scenarios from a human-robot team
performing an USAR task where each subteam $i$
consists of one human $H_i$ and one robot $R_i$.

\vspace{5pt}
\subsubsection{For subteam 1}
Based on the floor plan of the building in its search area, 
$R_1$ realizes that the team needs
to use an entrance to a hallway to start the exploration.
$R_1$ notices that a heavy object blocks the
entrance to the hallway.  Based on its
capability model of $H_1$ (i.e., what $H_1$ can and cannot lift) and $H_1$'s
goal, $R_1$ decides to interrupt its current activity and move the block
out of the way. $H_1$ and $R_1$ then continue exploring different parts of
the area independently when $H_1$ discovered a victim and informs $R_1$.  $R_1$
understands that $H_1$ needs to get a medical kit to be able conduct triage on
this victim as soon as possible but knows that $H_1$ does not know where
a medical kit is located.  Since $R_1$ has a medical kit already, but
cannot deliver it due to other commitments, it places its medical
kit along the hallway that it expects $H_1$ to go through, and informs $H_1$
of the presence of the kit.

\vspace{5pt}
\subsubsection{For subteam 2}
Based on the floor plan of the building in its search area, 
$R_2$ finds that all the entrances are automatic doors that are controlled from the inside.
Since the connection cannot be established due to power lost, 
the team needs to break a door open first. 
$R_2$ infers that $H_2$ is about to break a door
open based on the teaming context and its observations.  Since it
knows that breaking the door open may cause a board to fall on $H_2$, $R_2$
moves to catch the board preventatively.  Once $H_2$ and $R_2$ are inside,
however, $H_2$ is uncertain about the structural integrity and has no
information on which parts may easily collapse.  $R_2$ has access to the
building structure information and proposes a plan to split the search
in a way that minimizes human risk.

\vspace{5pt}
\subsubsection{For both subteams}
As both teams are searching their areas, they
receive information about a third area to be explored.  Since neither
$H_1$ nor $H_2$ are finished with their current search task, they assume
that the other will take care of the third area.  Since $R_1$ understands
$H_1$ and $H_2$'s current situation, and expects itself to be done with its
part of the task soon, $R_1$ decides to work on the third area since it
does not expect $H_1$ to need any help.  $R_1$ informs $H_1$.  $H_1$ is OK with it
and informs $H_2$ that team $1$ is working on the third area.  When $R_1$
arrives at the third area, it notices new situations which require
certain equipment from team $2$.  $R_1$ communicates with $R_2$ about the
availability of the missing items.  $R_2$ quickly predicts equipment
needs and anticipates that those items are not needed for a while.
After getting the OK from $H_2$ to lend the equipment to $R_1$, $R_2$
drives off to meet $R_1$ half-way, hand over the equipment, and $R_1$
returns to the third area with the newly acquired equipment.  $H_1$ was
not informed during this process since $R_1$ understands that $H_1$ has a
high workload.  Once the equipment is no longer needed, $R_1$ meets up
with $R_2$ again, returning the equipment in time for use by $H_2$.

\vspace{5pt}
Based on the above scenario, we can see that the mental modeling of the others
on a cognitive robotic teammate is critical to the fluent operation of the team
%
For example, $R_1$ needs to understand the capabilities of
$H_1$ (i.e., what $H_1$ can and cannot lift); both $R_1$ and $R_2$ need to be able
to infer about the intention of the human teammates.  The
modeling may also include the human's knowledge, belief, mental workload, trust, etc.
This human mental modeling for cognitive teaming between humans and robots 
connect with the three capabilities we introduced in Section \ref{sec:intro} as critical to 
the functioning of human-human teams and form the basis of the updated agent architecture in Fig.~\ref{fig:newagent} as follows - 

\vspace{5pt}
\textbf{C1. Recognizing teaming context to identify the status of the team task and states of the teammates}: 
For example, based on the floor plan of the building, 
$R_1$ realizes that the team needs
to use an entrance to a hallway to start the exploration.
$R_2$ finds that all the entrances are automatic doors that are controlled from the inside.
Consequently, it infers that the team needs to break a door open first. 
This inference process takes into account the modeling of the teammate's state 
(e.g., the intention to enter the building).

\vspace{5pt}
\textbf{C2. Anticipate team behavior under the current context}: 
For example, given that a heavy object blocks the
entrance to the hallway, $R_1$ infers that the human will be finding a way to clear the object. 
$R_2$ infers that $H_2$
is going to break a door open based on the teaming context and its observations.  
This prediction
takes into account of the modeling of the human's capabilities 
and knowledge about the teaming context.

\vspace{5pt}
\textbf{C3. Take proper actions to advance the team goal while taking into account the teammates}: 
For example, after anticipating the human's
plan, the robots should proactively help
the humans (e.g., $R_1$ helps $H_1$ move the block away and $R_2$ catches the
board preventively that can potentially hurt the human), while taking the account
the modeling of the human's capabilities, mental workload,
and expectation.

\vspace{5pt}
\paragraph{\textbf{Remark}} 
C3 above not only includes actions that contribute to the team goal,
but also actions for maintaining teaming situation awareness (e.g., making explanations).
As such, C3 feeds back to C1 and 
the three capabilities in turn form a loop that should be constantly exercised to achieve fluent teaming. 
Furthermore, although we have been focusing on implicit communication
(e.g., through observing behaviors) to emphasize the importance of mental modeling, 
explicit communication (e.g., using natural language) is also an important part of the loop. 
Another note is that since both implicit and explicit communication 
can update the modeling of the other teammates' mental states as discussed,
they are anticipated to evolve the teaming process in the long term.


\section{Challenges} 
\label{sec:challenges}

The capabilities reflected in the updated agent architecture present several challenges for the design of cognitive robotic teammates -- at the core of these issues is the need for an autonomous agent to consider not only its own model but also the human teammate's mental model in its deliberative process.  
In the following discussion, we will outline a few of our recent works in this direction, outline processes by which the agent can deal with such models, and end with a discussion on our work on learning and evaluating these models.


\begin{figure}[tbp]
\centering
\includegraphics[width=\columnwidth]{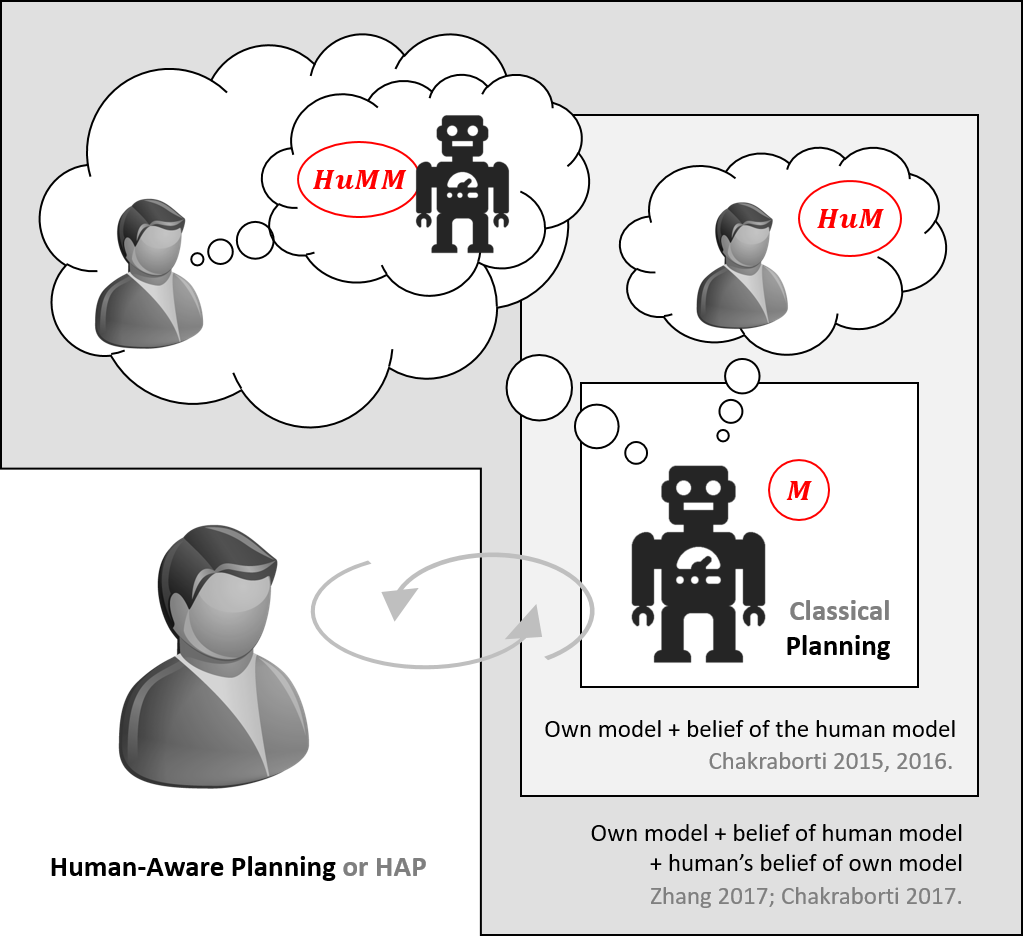}
\caption{Figure \cite{exp-fss} illustrating the expanding scope of the ``human-aware'' decision making process of an autonomous agent to account for the human model (HuM) and the human mental model (HuMM).}
\label{fig:hap}
\end{figure}

\subsection{Human Aware Planning}

Most traditional approaches to planning focus on one-shot planning in closed worlds given complete domain models.  While even this problem is quite challenging, and significant strides have been made in taming its combinatorics, planners for robots in human-robot teaming scenarios require the ability to be {\it human-aware}.  
We postulate then a departure from traditional notions of automated planning to account for with humans -- many of the challenges that arise from it are summarized in our work \cite{explain} under the umbrella of {\bf ``Multi-Model Planning''}. 
The term alludes to the fact that a cognitive agent, in course of its deliberative process, must now consider not only the model \textcolor{red}{$M$} that it has on its own but also the model \textcolor{red}{$HuM$} of the human in the loop, including the (often misaligned) {\em mental} model \textcolor{red}{$HuMM$} of the same task that the human might have. 
This setting is illustrated in Figure \ref{fig:hap}.
Here, by the term ``model'' of an robot, we include its action or {\em domain model} as well as its state information or {\em beliefs} and its goals or {\em intentions}. 
The human model allows the robot to account for the human's participation in the consumption of a plan, while the human mental model enables the robot to anticipate how the the plan will be {\em perceived} by the human as well as interactions that arise thereof. 

\begin{figure}[tbp]
\centering
\includegraphics[width=\columnwidth]{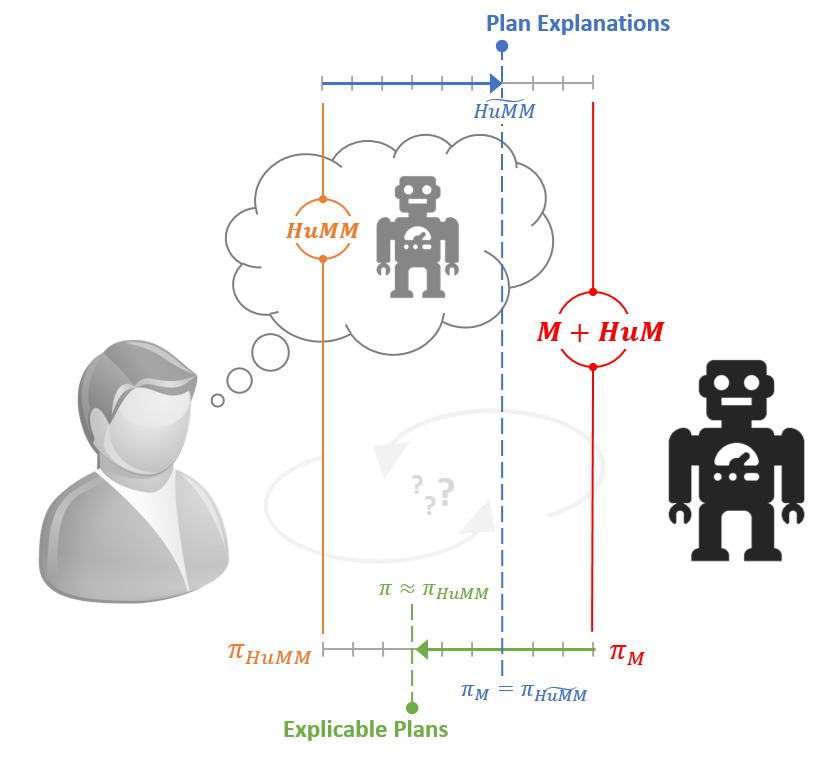}
\caption{Figure \cite{exp-fss} illustrating explanation generation \cite{explain} via the model reconciliation process, and explicable plan generation \cite{exp-yu} by sacrificing optimality in the robot's own model.}
\label{fig:mmp}
\end{figure}

\begin{figure*}
\centering
\includegraphics[width=\textwidth]{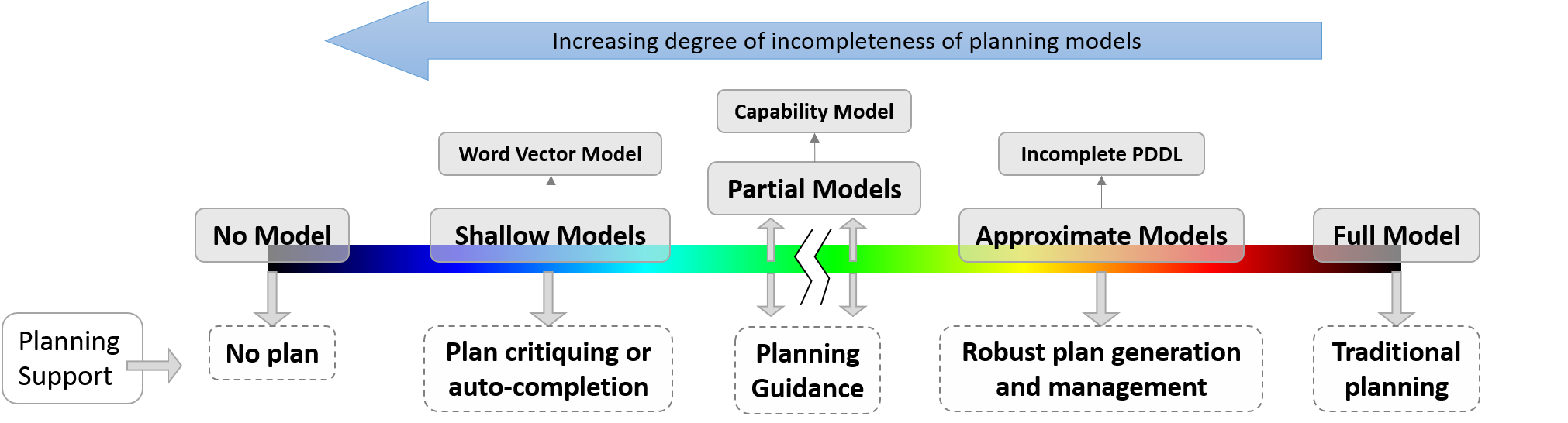}
\caption{Figure \cite{hankz-wv-reg} showing a schematic view of different classes of incomplete models and relationships between them in the spectrum of incompleteness.}
\label{fig:incomplete}
\end{figure*}

\subsubsection{Human-Robot Teaming / Cohabitation}

Incorporation of the human model \textcolor{red}{$HuM$} in the planning process allows the robot to take into consideration possible human participation in the task and thus identify its appropriate role in it.
This can relevant both when the robot is explicitly {\em teaming} \cite{kartik-iros-2014, narayanan2015automated, zhang2015human, talamadupula2017architectural}
with the human, or when it is just sharing or {\em cohabiting} \cite{tathagata-serendipity-iros,tathagata-conflict-aamas,mipc,Chakraborti2016AFF} the same workspace withouts shared goals and commitments. 
We have explored the typical roles of the robot in each of these scenarios -- e.g. in {\em planning for serendipity} \cite{tathagata-serendipity-iros} and in {\em planning with resource conflicts} \cite{tathagata-conflict-aamas} we looked at how a robot can plan for passive coordination with minimal prior communication, while in \cite{narayanan2015automated,zhang2015human} we explored the effects of {\em proactive support} on the human teammate. 
Indeed, much of existing literature on human-aware planning \cite{alami2006toward,alami2014human,tist,grandpa,tomic,cirillo2010planning,tathagata-serendipity-iros,tathagata-serendipity-iros,mipc} has focused on this setting; we will now explore additional challenges to the human-aware planning problem in the context of the human {\em mental} model \textcolor{red}{$HuMM$}.

\subsubsection{Explicable Task Planning}

One immediate effect of model differences between the robot and the human is that a robot even when optimal on its own can be suboptimal, and hence inexplicable, in the model of the human. 
This situation is illustrated in Figure~\ref{fig:mmp}.
When faced with such a situation the robot can choose to produce plans \textcolor{red}{$\pi$} that are likely to be more comprehensible to the human by being closer to the human expectations \textcolor{red}{$\pi_{HuMM}$}. 
This is referred to as {\em explicable planning} -- here the robot thus sacrifices optimality in its own model in order to produce more human-aware plans. 
There exists some recent work on {\em motion} planning while 
considering the human expectation both while computing trajectories \cite{Dragan-RSS-13,Dragan2015a,Dragan2015b}.
In recent work \cite{exp-yu,exp-anagha,zhang2016plan} we have explored how this can be achieved in the context of {\em task} both when the human model is perfectly known, or when it has to be learned in the course of interactions.
The latter work introduces a plan explicability measure \cite{exp-yu} learned approximately from labeled plan traces as a proxy to the human model.
This captures the human's expectation of the robot, which can be
used by the robot to proactively choose, or directly incorporate into the planning process to generate, plans that are more comprehensible without significantly affecting its quality.

\subsubsection{Explanation Generation}

Such plans, of course, may not always be desirable -- e.g. if the plan expected by the human is too costly (or even unsafe or infeasible) in the robot's model. Then the robot can choose to be optimal (\textcolor{red}{$\pi_{M}$}) in itself, and explain \cite{explain,exp-fss-multi} its decisions to the human {\em in terms of the model differences}.  
This process of {\em model reconciliation} ensures that the human and the planner remain on the same page in course of prolonged interactions. 
At the end of the model reconciliation process, the optimal plan in the agent's model become optimal in the update human model \textcolor{red}{$\widehat{HuMM}$} as well, as shown in Figure~\ref{fig:mmp}.
The ability to explain itself is a crucial part of the design of a cognitive teammate, especially for developing trust and transparency among teammates.
We argue \cite{explain} that such explanations cannot be a soliloquy -- i.e. the planner must base its explanations on the human mental model.
This is usually an implicit assumption in the explanation generation process; e.g. imagine a teacher explaining to a student - this is done in a manner so that the student can make sense of the information in their model.

\subsubsection{Human-Aware Planning Revisited}
\label{subsubsec:haware}

Sometimes the process of explanation, i.e. the cost of communication overhead, might be too high. However, at the same time, for reasons we explained above, there might not be any explicable plans available either. An ideal middle ground then is to strike a balance between the explicable planning and explanations.
We attempt to do this by employing model-space search \cite{exp-fss} during the planning process.
From the perspective of design of autonomy, this has two important implications - (1) as mentioned before, an agent can now not only explain but also {\em plan in the multi-model setting} with the trade-off between compromise on its optimality and possible explanations in mind; and (2) the argumentation process is known \cite{argue} to be a crucial function of the reasoning capabilities of humans, and now by extension of autonomous agents as well, as a result of algorithms that incorporate the explanation generation process into the decision making process of an agent itself.

\begin{figure*}
\centering
\includegraphics[width=\textwidth]{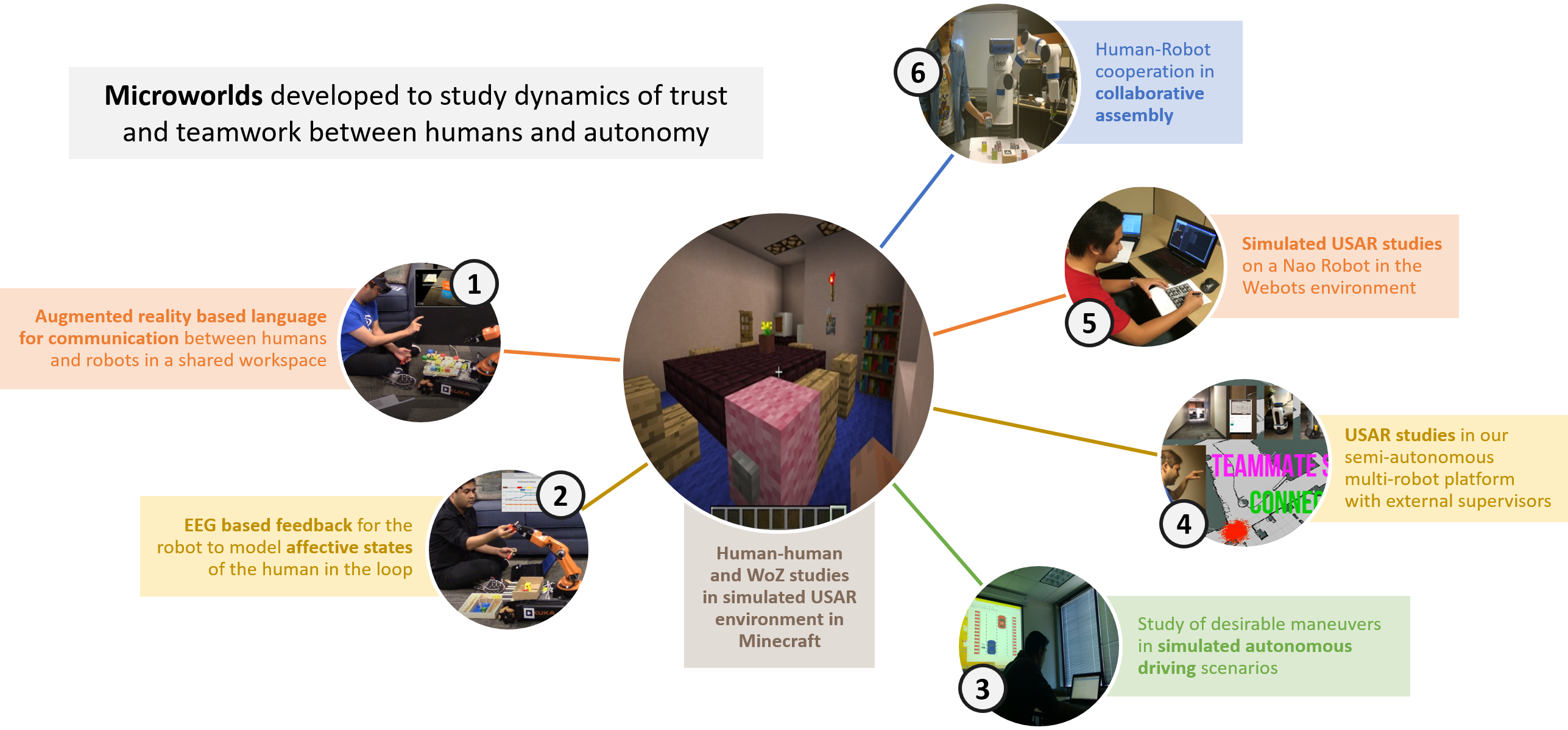}
\caption{Testbeds developed to study the dynamics of trust and teamwork between autonomous agents and their human teammates.}
\label{fig:testbeds}
\end{figure*}

\subsection{Learning Human (Mental) Models}

Of course, both the previous challenges were built on the premise that the human (mental) models are available or at least learned so as to facilitate the decision making process with these model in mind. Acquiring of such models, taken for granted among human teammates through centuries of evolution, is perhaps the hardest challenge to be overcome to realize truly cognitive teaming. The difficulty of this problem is exacerbated by the fact that much (specifically, $HuMM$) of these models cannot be learned from observations directly but only from continued interactions with the human. 

However, while much of the work on
planning has till now focused on complete world
models, most real-world scenarios, especially when they involve humans,
are {\em open-ended} in that planning agents typically do not have sufficient knowledge about all task-relevant information (e.g., human models) at planning time -- in
other words, the planning models would be incomplete.  
Despite being incomplete, such models must also support reasoning as well as be improvable from sensing, i.e. learnable.  
Hence, an important challenge is to develop representations of approximate and incomplete models that are easy to learn (for human mental modeling) and can support planning/decision-making (for anticipating human behavior).

Existing work on incomplete models (Figure \ref{fig:incomplete}) 
differ in the information that is available for model learning, as well
as how planning is performed.  
Some of them start with complete action
models and annotate them with
{\em possible} conditions to support incompleteness
\cite{tuan,hankz,hankz-macrops}.
Although these models support principled approaches for robust
planning, they are still quite difficult to learn.  
On the other end of the spectrum, are very shallow models \cite{hankz-wv-reg} that assume no structured information at all which are used mainly in short-term planning support such as action recommendation.  Partial models, that are somewhere in between, having more structured information while still being easy to learn \cite{yu-aamas} can prove to be powerful support for goal recognition. 
However, planning under such models is incomplete.  
In our work on explanation generation \cite{exp-fss-multi} we demonstrated how annotated models such as above can be used to deal with model uncertainty, while for explicable planning \cite{exp-yu,exp-anagha} we showed how CRF/Regression/LSTM-based models can be used to learn human preferences in terms of plan similarity metrics.

This is, however, only a start in this research direction. 
Performing human-aware planning with incomplete models remains an
important challenge in human-aware planning, especially given that we do not yet 
understand how these different human models interact.
For example, different human models capture different
aspects of the human (e.g., capabilities \cite{yu-aamas}, intentions \cite{6942970} and emotions \cite{scheutzschermerhorn09handbook})
which are closely inter-related, and it is not clear how they can be combined.  

\subsection{Communication and Evolution of Mental Models}

All these processes are aimed at bringing the robot's model and the human expectation of it closer over continued interactions. 
We have so far only discussed how a robot can maintain the human mental models. 
However, in teaming, this modeling is bi-directional. 
When robots do not have certain information, 
the robot can plan to sense and communicate with the human. 
In cases when the human is suspected to have insufficient
information about the robot, the robot needs to proactively communicate its model to the human.  
This communication can be, for example, about the
intention, plan, explanations and excuses of behavior \cite{excuses-icaps} for
the robot; or explanations that can not only make sense of the plan generation process itself \cite{kambhampati1990classification,sohBiaMcIAAAI2011,pat}, but must also make sense from the (robot's understanding of the) human's perspective \cite{explain}.
This is especially relevant in the case of human-robot teams, where the human's understanding of the robot may not be accurate.
Such explanations must not only be able to justify failure \cite{goebelbecker2010coming,Herzig:2014:RPT:3006652.3006726,menezes2012planning,eiter2010updating} but also the rationale behind a successful plan in order for the human to be able to reason about the situation and contrast among alternative hypotheses \cite{explain,lombrozo2012explanation,Lombrozo2006464}.

Further, explanations must be communicated at a level understandable to the human \cite{16roman-verbalization,16ijcai-verbalization}.
Communication can thus involve different modalities such as
visual projection, natural languages, gesture signaling, and a mixture of them \cite{DBLP:journals/corr/ChakrabortiSKK17}.  
This capability is important for the human and robot to evolve their mental models
to improve teaming in the long term.
Note that the models (i.e., the human actual model and its representation on the robot) are not required to be aligned, 
which is often only applicable for repetitive tasks \cite{Nikolaidis:2013}.
Much existing work on robots
communicating with humans using different modalities can be utilized \cite{kollaretal10hri, Tellex14askingfor}.  
However, a more critical challenge for the robot is
to compute when, what, and how
to communicate for model adaptation 
- communicating too much information can increase
the cognitive load of the human teammates while communicating too little can decrease the teaming situation awareness. 
Existing literature on decision support and human-in-the-loop planning \cite{radar,manikondaherding,chakraborti2017ubuntuworld} 
can provide insightful clues to dealing with such challenges in communication on information among teammates.

It must also be realized that
many human-robot teaming tasks are not only complex, 
but can also span multiple episodes for an extended period of time. 
In such scenarios, the system's performance is dependent on 
how the teams perform in the current task, 
as well as how they perform in future tasks.  
A prerequisite to consider long-term teaming is to maintain mental states of the agents (e.g., trust)
that influence their interactions, 
and analyze how these states dynamically affect the teaming performance and how they evolve over time. 

\subsection{Evaluation / Microworlds}

The design of human-machine systems is, of course, largely incomplete unless tested and validated in the proper settings.
Although the existing teamwork literature  \cite{cooke2013interactive} on human-human and human-animal teams has identified characteristics of effective teams --
in terms of shared mental models
\cite{Bowers,Mathieu2000}, team situational awareness
\cite{nancy-new}, and interaction \cite{nancy2013} --
it is less clear how much of those lessons carry over to human-{\em robot} teams.
To this end, we have developed a suite of testbeds or {\em microworlds} for generation and testing of hypothesis and rapid prototyping of solutions.
Figure \ref{fig:testbeds} illustrate some of the microworlds we have used so far -- anticlockwise from the left, this includes 
[1-2] a shared workspace \cite{DBLP:journals/corr/ChakrabortiSKK17} between humans and semi-autonomous agents that supports communication across various modalities such as speech, brainwaves (EEG) and augmented reality (AR) [video: \url{https://goo.gl/KdsEgr}]; [4-5] simulated Urban Search and Rescue scenarios \cite{narayanan2015automated,zhang2015human,exp-fss,exp-fss-multi} with internal semi-autonomous agents (humans and robots) supervised by external human teammates [video: \url{https://goo.gl/BKHnSZ}]; and [3,6] simulated (such as autonomous driving and collaborative assembly) domains for the study of multi-model planning \cite{exp-yu,exp-anagha,zhang2015human,narayanan2015automated} with humans in the loop [video: \url{https://goo.gl/UpEmzG}]. 
The aim here is to conduct human-human or Wizard of Oz studies \cite{nancy} in controlled settings and replicate desired behavior in the design of cognitive teammates.

\section{Conclusion}
\label{sec:conclusion}

In this paper, we discussed the challenges in design of autonomous robots 
that are cognizant of the cognitive aspects of working with human teammates.  
We argued that traditional goal-based and
behavior-based agent architectures are insufficient for building robotic teammates.
Starting with the traditional view of a goal-based agent, we expand it
to include a critical missing component: human mental modeling.  We
discussed the various tasks that are involved when such models are
present, along with the challenges that need to be addressed to
achieve these tasks.  We hope that this article can serve as guidance
for the development of robotic systems that can enable 
more natural teaming with humans.

\bibliographystyle{plain}
\bibliography{bib,cwm}

\end{document}